\documentclass[letterpaper, 10 pt, conference]{ieeeconf}
\IEEEoverridecommandlockouts

\usepackage{soul}
\usepackage{url}
\usepackage[hidelinks]{hyperref}
\usepackage[utf8]{inputenc}
\usepackage[small]{caption}
\usepackage{graphicx}
\usepackage{amsmath}
\usepackage{booktabs, longtable, multirow, array}
\usepackage[linesnumbered,ruled,vlined]{algorithm2e}
\usepackage[export]{adjustbox}
\usepackage{subfig}
\usepackage{subfig}
\usepackage[margin=.2cm]{caption}
\usepackage{algpseudocode}

\usepackage{etoolbox}
\usepackage[suppress]{color-edits}

\addauthor{lg}{blue}
\addauthor{hg}{orange}

\newcolumntype{M}[1]{>{\centering\arraybackslash}m{#1}}

\SetKwComment{Comment}{$\triangleright$\ }{}

\IEEEoverridecommandlockouts                              

\title{\LARGE \bf Federated Continual Learning for Socially Aware Robotics}

\author{Luke Guerdan$^{1*}$ and Hatice Gunes$^{2}$
\thanks{$^{1}$  School of Computer Science, Carnegie Mellon University, USA.
*Corresponding author: lguerdan@cs.cmu.edu. This work resulted from his ACS MPhil project at the University of Cambridge. 
$^{2}$ Department of Computer Science and Technology, University of Cambridge, UK.}}

\begin{document}

\maketitle


\begin{abstract}
From learning assistance to companionship, socially aware and socially assistive robotics is promising for enhancing many aspects of daily life. However, socially aware robots face many challenges preventing their widespread public adoption. Two such major challenges are (1) lack in behavior adaptation to new environments, contexts and users, and (2) insufficient capability for privacy protection. The commonly employed \emph{centralized learning} paradigm, whereby training data is gathered and centralized in a single location (i.e., machine / server) and the centralized entity trains and hosts the model, contributes to these limitations by preventing online learning of new experiences and requiring storage of privacy-sensitive data. In this work, we propose a \emph{decentralized learning} paradigm that aims to improve the personalization capability of social robots while also paving the way towards privacy preservation. First, we present a new framework by capitalising on two machine learning approaches, Federated Learning and Continual Learning, to capture interaction dynamics distributed physically across robots and temporally across repeated robot encounters. Second, we introduce four criteria (adaptation quality, adaptation time, knowledge sharing, and model overhead) that should be balanced within our decentralized robot learning framework. Third, we develop a new algorithm -- Elastic Transfer -- that leverages importance-based regularization to preserve relevant parameters across robots and interactions with multiple humans (users). We show that decentralized learning is a viable alternative to centralized learning in a proof-of-concept Socially-Aware Navigation domain, and demonstrate the efficacy of Elastic Transfer across our proposed evaluation criteria.
\end{abstract}




\section{Introduction}

As robot systems improve, they are moving from laboratory environments into our daily lives. The homes of tomorrow may enlist Jibo, ``the world's first family robot”, to lend support with cooking, family photos, and entertainment \cite{hodson2014first}. In the classroom, humanoid robots like Pepper may provide emotionally intelligent tutoring to students who need learning assistance \cite{pandey2018mass}. Yet, several key barriers remain before social robots such as these can become fully integrated into daily life. 

One barrier to widespread social robot adoption is \textit{personalization}. Whereas robots of today are designed for single encounters based on pre-configured abilities, future systems require adaptation over repeated long-term interactions \cite{IrfanRSPG-HRI22}. Pioneering studies in longitudinal HRI research have reported that robots who tailor their behavior towards users' personality, preferences, and interaction style foster more engagement and long-term acceptance \cite{leite2013social}. A further barrier to social robot adoption is the lack of \textit{privacy protections}. Because robots will support humans with sensitive activities such as learning and eldercare, data protection must be considered from the ground up. However, existing systems do not emphasize privacy, thus reducing user trust in the system and limiting widespread adoption \cite{butler2015privacy}.

An underlying cause of these barriers is the \textit{centralized machine learning} (ML) process robots use to develop intelligent behaviors. Centralized ML works by (1) gathering user data in a single location such as a server and (2) training a model to perform a new function (i.e.\ classifying user emotions) using the full dataset. However, this process limits robots' ability to personalize to new users, because doing so would require uploading raw user data, retraining the model, and redistributing the model to all devices. Moreover, in a social robotics scenario, storing user data on a server imposes privacy vulnerabilities. Instead, by leveraging \textit{decentralized learning}, whereby robots leave raw data on robot devices and learn by collaboratively exchanging model information, it may be possible to support personalization while also protecting privacy. 

In this work, we introduce a decentralized learning approach for socially aware robotics (\emph{FCL4SR} henceforth) by capitalising on two machine learning approaches, Federated Learning (FL) and Continual Learning (CL) \cite{yoonFCL2020}, to model interactions that are distributed \emph{physically} across robots and \emph{temporally} across repeated interactions. We outline four criteria that should be satisfied for \emph{FCL4SR} settings, including adaptation quality, adaptation time, effective knowledge sharing, and minimal overhead. As a proof-of-concept, we compare a range of existing Federated Continual Learning (FCL) methods on a socially-assistive navigation benchmark with respect to these four criteria. We also develop a new regularization-based algorithm, \emph{Elastic Transfer}, that improves performance with respect to three of the proposed criteria. This new method improves performance by penalizing changes to parameters that were important in earlier local tasks as well as tasks encountered by other clients by unifying existing FL and CL approaches \cite{shoham2019fedcurv,kirkpatrick2017overcoming}. In summary, this work provides three key contributions by:

\begin{enumerate} 
    \item \textbf{Formulating socially aware robotics as a decentralized learning problem.} To our knowledge, this is the first work to formulate socially aware robotics as a decentralized learning problem that combines both FL and CL. We provide a flexible extension to previous CL for socially aware robotics settings (\emph{FCL4SR}) that supports \textit{personalization} to new individuals and groups.

    \item \textbf{Developing a decentralized learning evaluation framework.} We outline which tradeoffs should be balanced in \emph{FCL4SR} settings, and apply this framework to a comprehensive evaluation of 10 existing FCL approaches. 

    \item \textbf{Proposing a regularization strategy for improving FCL performance.} We draw connections between existing FL and CL methods to propose a new \textit{Elastic Transfer} approach for improving knowledge sharing among clients. This provides the first regularization-based strategy tailored to FCL and improves adaptation quality, adaptation time, and knowledge sharing.
\end{enumerate}

\section{Related work}\label{sec:related_work}

In \textbf{Continual Learning (CL)}, models learn online from sequentially encountered data \cite{parisi2019continual}. Given a series of incremental \textit{tasks} (e.g., interactions with new users, or repeated interaction sessions with the same user), CL approaches aim to incrementally learn each task without forgetting knowledge from prior tasks. \textit{Catastrophic forgetting} occurs if performance on prior tasks deteriorates while learning the current task. Several approaches have been proposed to address catastrophic forgetting in CL. Regularization-based strategies add a penalty to the loss term that prevents models from diverging from the one learned on prior tasks. L2-transfer (L2T) \cite{li2017learning} applies an L2 penalty, while Elastic Weight Consolidation (EWC) \cite{kirkpatrick2017overcoming} applies a quadratic penalty based on Fisher information to consider parameter importance. Architectural CL strategies, such as Additive Parameter Decomposition (APD) \cite{yoon2019scalable}, address catastrophic forgetting by modifying the model structure to accommodate task specific information.

In \textbf{Federated Learning (FL)}, a network of distributed phones, robots, or other clients downloads a common initial model from a server, trains on locally available data, then transmits updated parameters to the server. The server aggregates updates into a new global model and re-distributes it to clients. FedAvg is a standard approach for aggregating client parameter updates on the server via a weighted average \cite{mcmahan2017communication}. However, prior work shows that FedAvg performance is poor in non-iid settings with client dataset size imbalances, concept drift (same label is assigned for dissimilar features), or concept shift (different label is assigned to similar features). FedProx handles this by imposing an L2 penalty in the client objective to prevent divergence from the most recent aggregate model \cite{li2018fedprox}. FedCurv extends FedProx by imposing an importance-based elastic penalty based on Fisher information \cite{shoham2019fedcurv}.

There are recent works on personalized federated learning that learn a global model while also fine tuning local models on user-specific information \cite{rudovic2021personalized,mansour2020three}. Personalized federated learning approaches have been reported to be vulnerable to \textit{negative transfer}, whereby averaging user-specific models worsens performance for some subjects \cite{rudovic2021personalized}. Existing personalized FL methods have also seen limited consideration of temporal adaptability, where concept shift and concept drift cause non-iid data over time \cite{tan2021towards}. In this work, we consider continual learning as the solution well-suited to address these issues. 

\begin{figure}[t!]
    \centering
    \includegraphics[width=\linewidth]{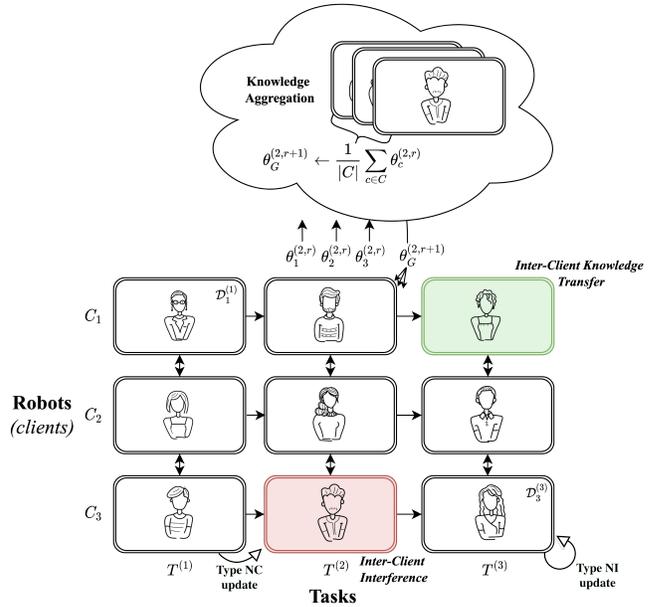}
    \caption[Schematic.]{A schematic diagram of Federated Continual Learning for Socially Aware Robots (\emph{FCL4SR}). In this example, a joint global model $\theta_G$ is learned by a network of three robots (i.e., \textit{clients}) over the course of interactions with three groups of participants. Each user's data $\mathcal{D}_{i}^{j}$ remains locally on the device, and only parameter updates are transmitted (federated learning). When robots interact with a new group of users, local data is erased from the device and robots attempt to retain knowledge from interactions with prior users (continual learning).}
    \label{fig:fcrl}
\end{figure}

In \textbf{Federated Continual Learning (FCL)}, FL and CL are combined to accommodate learning scenarios that are spatially and temporally distributed. A network of distributed clients each learns a series of local tasks, where each task may be available to a subset of other clients, or be available on a single client \cite{yoonFCL2020}. This paradigm introduces a new opportunity for \textit{inter-client knowledge transfer}, whereby experiences from previous tasks can be shared among clients to improve learning. However, it also introduces a challenge of \textit{inter-client interference}, in which experiences aggregated from other clients accelerate catastrophic forgetting.

FedWeIT is an architectural approach that was initially proposed to solve the FCL problem \cite{yoonFCL2020}. This technique decomposes a global model into a set of client-specific base parameters and task-adaptive parameters that learn knowledge about each local task. This approach encourages knowledge transfer by enabling clients to exchange task-adaptive parameters and weight them locally via an attention mechanism. Recent work has proposed new FCL methods for addressing inter-client interference and facilitating knowledge transfer \cite{ huang2022learn, criado2022non, wei2022knowledge, wang2023federated, estevez2021concept}. These approaches sometimes explicitly model learning as an FCL problem \cite{wang2023federated}, and other times examine FL under various forms of non-iid data \cite{huang2022learn, criado2022non, wei2022knowledge, estevez2021concept}. While FCL has been applied to natural language processing \cite{chaudhary2022federated} and healthcare \cite{lu2022personalized} applications, no work to our knowledge has explored applications of FCL to socially aware robotics. 

\textbf{Decentralized Learning in Socially Aware Robotics.} Continual learning \cite{lesrotCLRobots}, federated learning \cite{xianjia2021federated}, and a combination of both frameworks \cite{liu2019lifelong} have been applied to broader robotics problems. Continual learning has been shown to be an effective approach for personalization of socially-aware robots over repeated interactions \cite{IrfanRSPG-HRI22,logacjov2021learning}. Most closely related to this work, a recently-proposed framework extends an existing CL taxonomy \cite{lesrotCLRobots} to affective robotics. Their framework formulates continual learning of personalized socio-emotional human robot interactions  \cite{cl4ar}, and has been successfully leveraged for learning the social appropriateness of domestic robot actions \cite{TjomslandKG22}. A user study applying this framework demonstrated that CL-based personalization over an interaction session improves participants’ impressions of robot anthropomorphism and likeability \cite{churamani2022continual}.

While results demonstrate the feasibility of CL in human robot interaction, challenges remain. For instance, strategies based on continual learning only do not provide a mechanism for exchanging information among multiple robots. Therefore, performance will be limited by training data encountered previously on the local device. This concern is especially relevant in socially aware robotics and human robot interaction contexts, where dataset sizes tend to be comparably small \cite{cl4ar}. In this work, we propose leveraging FL to overcome such dataset and data sharing limitations, while still ensuring that learning remains \textit{distributed} (e.g., does not require centralized storage of raw data on a server). As a result, our approach also provides a promising step towards achieving better privacy-preservation.

\section{Approach}
In this section, we present a new framework for formulating decentralized socially aware robot learning as an FCL problem, \emph{FCL4SR}, and identify key desiderata that should be achieved by the \emph{FCL4SR} framework. Below we describe existing approaches before formulating a new regularization-based method called \textit{Elastic Transfer}. 

\subsection{Learning Scenario}\label{sec:learning_scenarios}

In Federated Continual Learning for Social Robots (FCL4SR), a network of social robots adaptively learns new intelligent behaviors over a sequence of distributed encounters with end-users (Fig. \ref{fig:fcrl}). For example, social robots may learn to detect user-specific nuances in affective signals \cite{cl4ar} or social navigation preferences \cite{manso2020socnav1, TjomslandKG22}. In FCL4SR scenarios, data used to learn new behaviors is partitioned into client-task datasets \lgedit{$\mathcal{D}_{i,j}$} containing data acquired from the $i$’th robot during an interaction with the $j$’th user. During each task, robots perform multiple rounds of local training and parameter aggregation to learn the desired behavior (FL step). The robots then begin interacting with a new set of end-users, and loose access to previously-acquired raw data (CL step). Thus, \emph{FCL4SR} only requires raw user data to be temporarily stored on the robot during the interaction with a current user, and does not require permanent storage on-device or on a remote server. As such \emph{FCL4SR} provides improved privacy protections compared to permanently storing raw user data on a server (i.e., as is standard in current centralized learning practices), but does \textit{not} provide a formal privacy guarantee (i.e., as would be provided by differential privacy).

Our \emph{FCL4SR} learning scenario builds upon a prior framework designed to describe learning personalized affective behaviors in CL. In the prior framework, \textit{Type NC} updates occur when a robot learns new affective behavior categories (i.e.\ learning to distinguish happiness vs.\ sadness facial expressions) over each task. Similarly, \textit{Type NI} updates occur when a robot learns variations between the same affective behavior (i.e.\ different variations of happiness facial expressions) \cite{cl4ar}. In our framework, we use Type NC updates to describe personalization to \textit{new users} and Type NI updates to describe repeated interactions with the \textit{same user}. Further, interactions with new users can be distributed spatially (i.e., when multiple robots interact with different users and share information via FL) and temporally (i.e., when the same robot interacts with different users over time). Fig. \ref{fig:fcrl} shows a schematic illustration of this learning scenario. 

As a concrete example of \emph{FCL4SR}, consider a hospital where a network of intelligent robots is deployed for emotional support [23]. In this setting, robots learn appropriate behaviors over repeated interactions with a patient. Because users may have different response patterns, it is desirable to update the model based on individualized data (i.e. personalization). Given the sensitive setting, it is also critical to limit sharing of sensitive raw-data via a server or persistent local storage on-device. In FCRL, each robot gathers data during interactions with a patient, trains on this data locally, and transmits parameter updates for updating a global model. When the robots interact with a new set of patients, local data is removed, and the information encoded in learned parameters remains available during interactions with future residents. Thus, an FCRL algorithm exchanges information among robots to \lgedit{accelerate learning of intelligent behaviors with new groups of end-users.} \lgdelete{on future tasks. If parameters learned from a previous patient accelerate adaptation for a future patient interacting with a different robot, we say that inter-client knowledge transfer occurred. Yet, if parameters learned from a new patient override this information we say that inter-client interference occurred.} \lgedit{While the example above focuses on a medical setting, our proposed framework also applies to diverse settings such as tutoring \cite{alves2019empathic}, well-being coaching \cite{churamani2022continual}, autism therapy \cite{rudovic2018personalized}, where social robots adaptively learn new intelligent behaviors from end-users.}

\subsection{Evaluation Framework}

It is crucial to examine several performance dimensions to gain a holistic understanding of FCL tradeoffs. For example, one naive approach for solving the scenarios above would be to train a separate model for each client-task pairing and load the model dynamically during test time. This Single Task Learning (STL) method may provide strong performance (high \textit{adaptation quality}). However, it would also require many feedback rounds to provide sufficient data. 

To improve the \textit{adaptation time}, we may enhance STL by running a local CL algorithm on each robot (termed Local-CL). Here, the initial set of participants still provide many feedback examples, but later participants benefit from fine-tuning an initialized model. However, if the first participants return to interact with the robot in later tasks, the robot may require additional interactions to recover the initial adaptation quality due to \textit{catastrophic forgetting}. A further drawback of CL and STL is that they are unable to leverage experiences from other robots. Therefore, to facilitate \textit{knowledge sharing} among robots, we ca share parameter updates through FL. Thus, an approach using both CL and FL (FL-CL) such as FedWeIT can be used. 

Other trade-offs to consider include \textit{privacy} and \textit{model overhead}. Because STL, Local-CL, and FL-CL employ decentralized learning, they all improve user privacy. However, it is important to consider overhead factors including model size, client-server communication costs, and energy consumption \cite{lesrotCLRobots}. We summarize these considerations in the following desiderata:

\begin{enumerate}
  \item \textbf{High Adaptation Quality.} After a series of interactions, the robot should be well-tuned to the specific context. This will be demonstrated by satisfying the learning objective (i.e. low MSE, high F1-score) at the end of training.
  
  \item \textbf{Minimal Adaptation Time.} The robot should quickly adjust to the interaction context. This will be achieved by (1) generalizing to new predictive tasks with limited \lgedit{fine-tuning} and (2) learning effectively on small datasets. 
 
  \item \textbf{Effective Knowledge Sharing.} The robot should integrate knowledge from other robots to achieve high adaptation quality with fast adaptation time.

  \item \textbf{Minimal Overhead.} The robot's financial, energy, and connectivity requirements should be minimized. 
\end{enumerate}

\subsection{Elastic Transfer}

Based on an evaluation of existing FCL approaches (exp.\ one and two below), we formulate a new regularization-based method. Our approach extends EWC in the CL setting and FedCurv in the FL setting. Both EWC and FedCurv are motivated by a Bayesian framework. In a CL scenario where an optimal model $\theta^{1}_{*}$ has been found for Task 1, EWC finds a solution for Task 2 near $\theta^{2}_{*}$ while remaining within the region of low Task 1 error. This is achieved by applying a quadratic penalty to each parameter proportional to its importance in previous tasks \cite{kirkpatrick2017overcoming}.

The process of optimizing the parameters $\theta$ given task datasets $\mathcal{D}^{1:T} = \{\mathcal{D}^{1}, \mathcal{D}^{2}, ..., \mathcal{D}^{T} \}$ can be framed as computing the posterior $p\left(\theta \mid \mathcal{D}^{1}, \ldots, \mathcal{D}^{T}\right)$ over all possible values of $\theta$. As the model learns a new task online, it combines the prior learned from earlier tasks with the $t$'th task's likelihood to form an updated posterior given by: 

 \footnotesize
\begin{equation} \label{eq:ewc_bayes}
\begin{split}
    \frac{p\left(\mathcal{D}^{t} \mid \theta\right) p\left(\theta \mid \mathcal{D}^{1}, \ldots, \mathcal{D}^{t-1}\right)}{p\left(\mathcal{D}^{t} \mid \mathcal{D}^{1}, \ldots, \mathcal{D}^{t-1}\right)}.
\end{split}
\end{equation}
\normalsize

In this case, only $\mathcal{D}^{t}$ is required while learning task $t$ because the posterior absorbs knowledge from previous tasks. EWC makes use of Laplace's approximation to estimate this intractable posterior using a Gaussian \cite{mackay1992practical}. In particular, given a mean $\theta^{i}_{*}$ centered at the maximum a posteriori (MAP) estimate of task $i < t$ and a precision defined by the diagonal Fisher information matrix (FIM) $F^{i}$ evaluated at $\theta^{i}_{*}$, Laplace's approximation can be given by $p(\mathcal{D}^{i} \mid \theta) \approx \mathcal{N}(\theta ; \theta^{i}_{*}, F^{i})$. Provided that $\theta^{i}_{*}$ is a local minima, the FIM acts as a surrogate to the Hessian of the negative log-likelihood of the posterior distribution \cite{huszar2017quadratic}. Given this approximation, the MAP estimate $\theta^t_{*}$ for task $t$ can be found by minimizing: 

\small
\begin{equation} \label{eq:fisher_loss}
    -\log p\left(\mathcal{D}^{t} \mid \theta\right) +\frac{\lambda}{2} \sum_{i=0}^{t-1}F^{i}\left(\theta-\theta_{*}^{i}\right)^{2} 
\end{equation}
\normalsize

where $\lambda$ controls regularization strength.
FedCurv applies this same posterior decomposition over client devices in an FL setting \cite{shoham2019fedcurv}. During each communication round $r$, each client $c$ exchanges their MAP estimate $\hat{\theta}_{c,r}$, and precision estimate $F_{c, r}$ refined over $E$ epochs local training. The MAP estimate for client $c$'s parameters at round $r$ of training can be given by minimizing:  

\small
\begin{equation} \label{eq:fedcurv_loss}
     -\log p\left(\mathcal{D}_{c} \mid \theta\right)+\frac{\lambda}{2} \sum_{i \in C \setminus{c}}F_{i, r-1}\left(\theta-\hat{\theta}_{i, r-1}\right)^{2}
\end{equation}
\normalsize

where the second term provides an importance-based regularization penalty proportional to $\lambda$. Because each client trains for $E$ epochs-per-round rather than to task convergence, the MAP estimates $\hat{\theta}_{c, r}$ maybe not be at a local minima, and approximates the true MAP. This can reduce FedCurv performance with small $E$ \cite{shoham2019fedcurv}. A key benefit of EWC and FedCurv is the adaptive penalty term imposed on the training objective. Whereas FedProx and L2T impose isotropic L2 penalties that constrain parameters equally, importance-based regularization penalizes changes to parameters in proportion to their salience (e.g. elastic regularization). \lgdelete{With EWC, this prevents overwriting knowledge from earlier tasks. With FedCurv, this prevents convergence instabilities introduced by non-iid data across clients.} 

We extend elastic regularization to improve FCL convergence to be used in socially aware robotics settings. By constraining learning on local client-task datasets such that important parameters are not overwritten across \textit{clients and earlier tasks}, it may be possible to limit inter-client interference. Similarly, this process may also promote consolidation of salient parameters learned across clients, promoting effective knowledge sharing. Whereas FedWeIT proposes transfer of task-adaptive \textit{parameters} between clients, we instead propose transfer of task-adaptive \textit{importance estimates} encoded by $\theta*$ and $F*$.

\begin{figure}
    \centering
    \includegraphics[width=.8\linewidth]{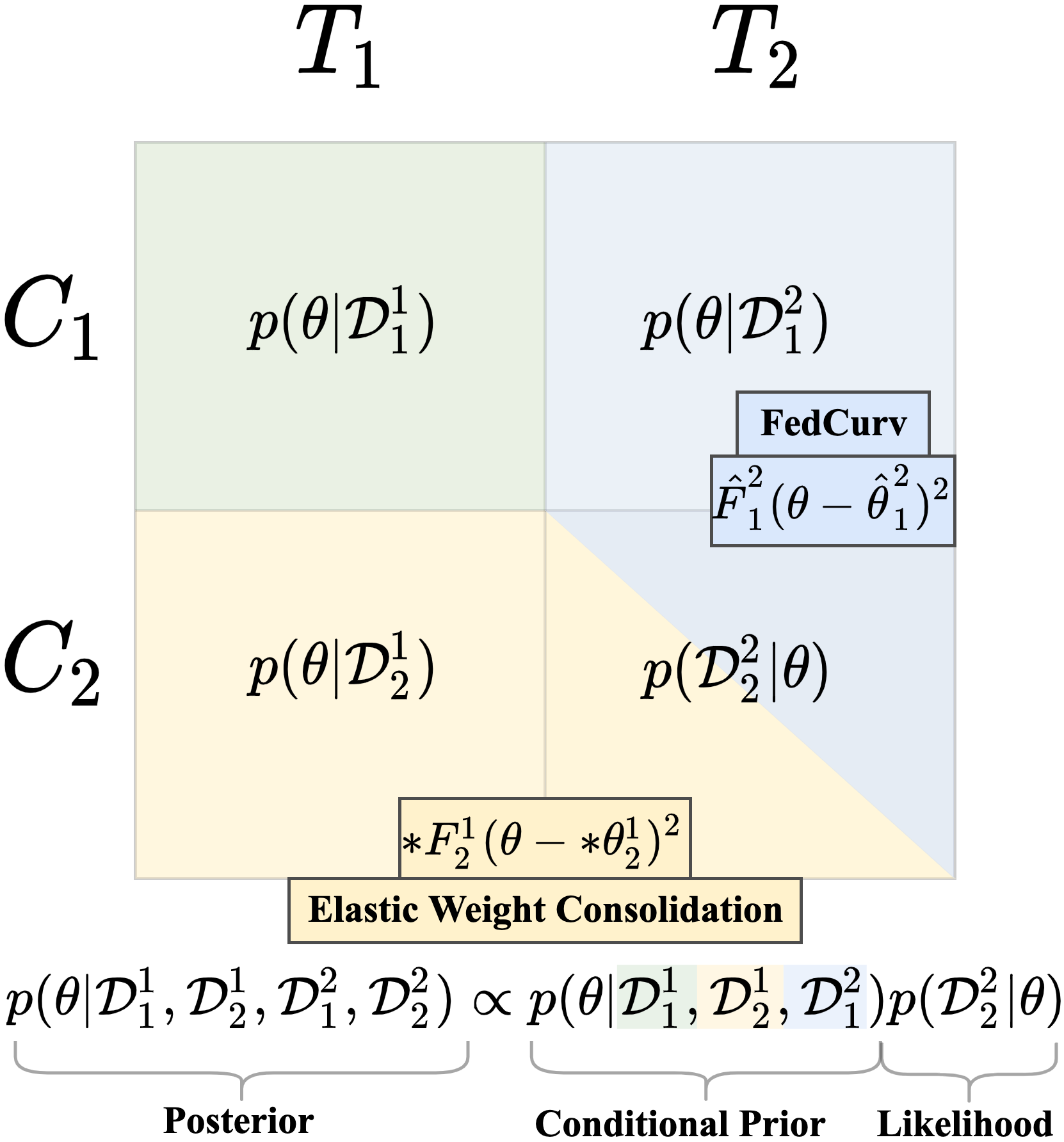}
    \caption[Elastic Transfer training schematic diagram.]{Training on the local dataset $\mathcal{D}_{2}^{2}$.}
    \label{fig:elastic_transfer}
\end{figure}

Fig.\ \ref{fig:elastic_transfer} summarizes how FedCurv and EWC can be extended to FCL problems. This schematic illustrates the objective of training a local model on $\mathcal{D}_{2}^{2}$ in a scenario with two clients and two tasks. We formulate the process of learning $\mathcal{D}_{2}^{2}$ as incorporating the likelihood into the conditional prior of other client-task partitions to give a new posterior. We approximate this term by combining three cases: 

\begin{itemize}

\item \textbf{Case 1: Previous task, same client}. Here, we approximate $p\left(\theta\mid\mathcal{D}_{2}^{1}\right)$ using $*\theta_{2}^{1}$, $*F_{2}^{1}$ (EWC; beige in Fig.\ \ref{fig:elastic_transfer}). Because $*\theta_{2}^{1}$ is a previous task trained to convergence,  we expect $*\theta_{2}^{1}$ to be at a local minima. 

\item \textbf{Case 2: Previous task, different client}. Here (shown in green), $p\left(\theta\mid\mathcal{D}_{1}^{1}\right)$ is an earlier task learned on client 1. To prevent overwriting knowledge from $\mathcal{D}_{1}^{1}$ while learning $\mathcal{D}_{2}^{2}$, we also need to account for $*\theta_{1}^{1}$, $*F_{1}^{1}$ in the local objective.

\item \textbf{Case 3: Same task, different client}. In this case (FedCurv; shown in blue), when we approximate $p\left(\theta\mid\mathcal{D}_{1}^{2}\right)$, we use $\hat{\theta}_{1}^{2}$, $\hat{F}_{1}^{2}$ from other clients as an MAP and precision estimate. Because task 2 is also being trained on other clients, $\hat{\theta}_{1}^{2}$ acts as a rough approximation of the MAP where $\hat{\theta}_{1}^{2}\approx*\theta_{1}^{2}$ for large $E$.

\end{itemize}

We leverage Online-EWC to scale the scenario shown in Fig.\ \ref{fig:elastic_transfer} to a set of clients $C$, each of which learn a series of $T$ tasks. Online-EWC applies Laplace's approximation to the full posterior rather than each term individually by storing a single FIM estimate $F^{*}_{c}$ and re-centered MAP estimate $*\theta_{c}^{t-1}$ \cite{schwarz2018progress}. The online FIM estimate $F_{c}^{*} := {\sum_{k=1}^{t-1} *F_{c}^{k}}$ is computed as a running sum of Fisher information matrices estimated for each of the the previous $t-1$ tasks on client $c$. The MAP estimate $*\theta_{c,r}^{t}$ in the $r$'th round of training can be found my minimizing:
\small
\begin{equation}\label{eq:et}
\begin{split}
     -\log p\left(\mathcal{D}_{c,t} \mid \theta\right)+
    \frac{\lambda_1}{2}\sum_{i \in C }*F_{i}(\theta-*\theta_{i}^{t-1})^2\\
    +\frac{\lambda_2}{2}\sum_{i \in C \setminus{c}}\hat{F}_i(\theta-\hat{\theta}_{i,r-1}^{t})^2.
\end{split}
\end{equation}
\normalsize
The first term uses $\lambda_1$ to weight \textit{refined} estimates $*F$, $*\theta$ from previous tasks, while the second uses $\lambda_2$ to weight \textit{rough} estimates of $\hat{\theta}$, $\hat{F}$ currently being optimized across clients. The first term applies case 1 when $c=i$, and case 3 when $c \neq i$. The second term handles case 2. We show the training process for optimizing this local loss in Alg. \ref{alg:elastic_transfer}. Under this regime, storage requirements are linear in $|C|$, while communication requirements are quadratic in $|C|$.

\begin{algorithm}[t]

\caption{{\sc Elastic Transfer}}\label{alg:elastic_transfer}
    \DontPrintSemicolon
    \KwIn{Client-task datasets $\{\mathcal{D}_{c}^{t}\}_{c \in C, t \in \{1,...,T\}}$, epochs-per-round $E$, rounds-per-task $R$,  Clients $C$, Number of tasks $T$}
    \KwOut{Optimized global model $\theta_G$}
    
    \For{$t \gets 1$ \textbf{to} $T$} {
        \uIf{$t > 1$}{ 
            \For{$c \in C$} {
                Update $*F_c$, $*\theta_{c}^{t-1}$ from previous task\;
                Send $*F_c$, $*\theta_{c}^{t-1}$ to other clients
            }
        }
        \For{$r \gets 1$ \textbf{to} $R$} {
            Select available clients $C_{r} \subseteq C$\;
            \For{$c \in C_{r}$} {
                $\hat{\theta}_{c}^{t}, \; \hat{F}_{c}^{t} \gets TrainLocal(\theta_G, E)$
                \Comment*[r]{minimize  e.q.\ \ref{eq:et} on device}
                Send $\hat{\theta}_{c}^{t},\; \hat{F}_{c}^{t}$ to other clients 
            }
            $\theta_G \gets \sum_{\substack{c \in C_r}} \frac{1}{|C_r|} \hat{\theta}_{c,t}$
        }
    }
    \Return{$\theta_G$}
    \label{algo:etr}
\end{algorithm}

\begin{figure}[t]
    \centering
    \includegraphics[width=.7\linewidth]{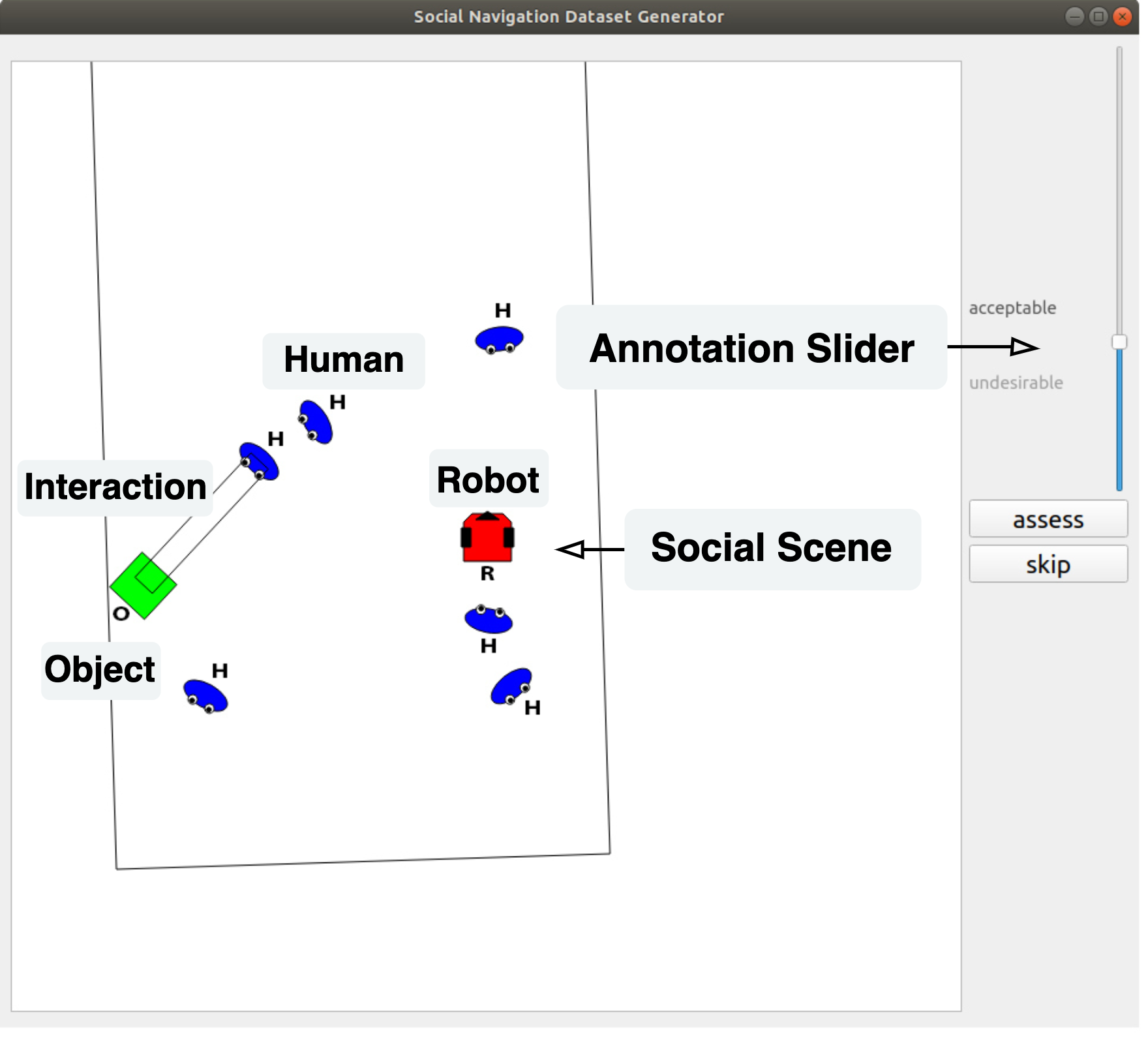}
    \caption[Scenario.]{Experimental setup for collection of SocNav1 data \cite{manso2020socnav1}}.
    \label{fig:annotation_task}
    \vspace{-5mm}
\end{figure}

\section{Evaluation}\label{sec:setup}

To demonstrate the utility of FCL4SR in socially aware robotics scenarios, we conduct a case study using a Socially-Aware Navigation (SAN) task. In SAN tasks, robots learn to position themselves appropriately around humans based on social conventions. SAN tasks provide a strong application setting for FCL4SR because humans have differing preferences regarding robot navigation patterns \cite{rios2015proxemics}. These preferences can vary across users, as well as within users depending on the social context \cite{syrdal2007personalized}. FCL4SR provides a method for continuously adapting robots' understanding of user navigation preferences over multiple distributed interactions without requiring centralized storage of user data. 

\subsection{Dataset}
 We use the SocNav1 benchmark for evaluating socially-aware robot navigation models \cite{manso2020socnav1}. This dataset contains 9,280 annotated examples gathered from 12 participants using a desktop-based tool. Participants were presented a static social scenario including humans, robots and objects, and were instructed to \textit{``assess the robot’s behavior in terms of the disturbance caused to humans"} on a scale from 0 to 100 \lgedit{(Figure \ref{fig:annotation_task})}. See \cite{manso2020socnav1} for additional task details.

\subsection{Centralized Baseline}
The SocNav1 authors developed a graph neural network (GNN) architecture for learning SocNav1 scenes \cite{manso2020graph}. We replicated their GNN comparison to determine suitable hyperparameters and establish a centralized learning baseline assuming standard train/validation/test partition on a server.  We vary the number of network layers, hidden layer size, weight decay ($\gamma$), and mini-batch size. A Graph Convolutional Network implemented via Deep Graph Library (GCN+DGL) with 8 layers and 32 hidden units per layer, $\alpha=5e^{-3}$, $\gamma=1e^{-4}$, and a mini-batch size of 32 performed the best (test MSE=.0296). We use this configuration in all following experiments. 

\subsection{FCL4SR Experimental Setup}

We construct an FCRL scenario with three clients and four tasks (matching Fig. \ref{fig:fcrl}) using the SocNav1 dataset. Specifically, we assign annotations from each of the 12 SocNav1 participants to a distinct client-task dataset such that each robot personalizes to the spacial navigation preferences of a new participant during each task.  We split each data into a 70/15/15 train, validation, test split (Table \ref{table:scenario_b}).

Because SocNav1 participants demonstrated inter-rater variability on a commonly-rated subset of scenes \cite{manso2020socnav1}, this scheme mirrors inter-subject variability present in real-world human-robot interaction scenarios. We construct a scenario with three clients and four tasks (as opposed to other possible factorizations of 12 client-task datasets such as 2 clients, 6 tasks etc.) because it provides a reasonable trade-off between requiring long-term retention of knowledge over four users (i.e., tasks) and parameter sharing among three clients. We augment data via the same procedure employed by SocNav1 authors, which involves mirroring social scenarios over the vertical axis \cite{manso2020socnav1}. All experiments and results reported in this paper are on the held out test set with data augmentation. 

\begin{table}[t]
\centering
\resizebox{\columnwidth}{!}{%
        \begin{tabular}{ll|llll}
        \toprule
                          &          &  \textbf{Task 1} &  \textbf{Task 2} &          \textbf{Task 3} & \textbf{Task 4} \\
        \midrule
        \multirow{3}{*}{\textbf{}} & \textbf{Client 1} &     111, 24, 24 &   781, 168, 168 &     417, 90, 90 &   86, 19, 19 \\
                          & \textbf{Client 2} &      86, 18, 19 &     365, 78, 79 &  1750, 375, 375 &  431, 92, 93 \\
                          & \textbf{Client 3} &  1750, 375, 375 &     103, 22, 23 &   46, 10, 10 & 565, 121, 122  \\
        \cline{1-6}
        \multirow{3}{*}{\textbf{+ Aug.}} & \textbf{Client 1} &     222, 24, 24 &  1562, 168, 168 &     834, 90, 90 &  172, 19, 19 \\
                          & \textbf{Client 2} &     172, 18, 19 &     730, 78, 79 &  3500, 375, 375 &  862, 92, 93 \\
                          & \textbf{Client 3} &  3500, 375, 375 &     206, 22, 23 &  92, 10, 10 & 1130, 121, 122   \\
        \bottomrule
        \end{tabular}
     }  

    \caption[Benchmark Client-task dataset sizes]{Client-task train, validation, test dataset sizes with and without data augmentation. } \label{table:scenario_b}
\end{table}

\subsubsection{Evaluation Measures}
We compute AMSE, BWT, and FWT as specified in \cite{lesrotCLRobots}. We compute the training-test performance matrix $P$, where rows denote training datasets $Tr^{(i)}$ available to the algorithm during each task, and columns denote the test datasets $Tr^{(j)}$. We compute AMSE over tasks as $\frac{1}{T}\sum_{i=1}^{T} P_{i, i}$.We compute BWT by measuring the performance difference between the previous task and the current task
\begin{equation}\label{eq:bwt}
    BWT = \frac{\sum_{i=2}^{N} \sum_{j=1}^{i-1}\left(P_{i, j}-P_{j, j}\right)}{\frac{N(N-1)}{2}}.
\end{equation}

Because we measure BWT in the context of MSE, negative BWT indicates performance is improving over tasks. We compute FWT via the mean of the \lgedit{future unseen tasks}\lgdelete{elements in Fig.\ \ref{fig:P_matrix}} such that

\begin{equation}\label{eq:fwt}
    FWT = \frac{\sum_{i=1}^{j-1} \sum_{j=1}^{N} P_{i, j}}{\frac{N(N-1)}{2}}.
\end{equation}

\begin{figure}
\centering
    \includegraphics[width=\linewidth]{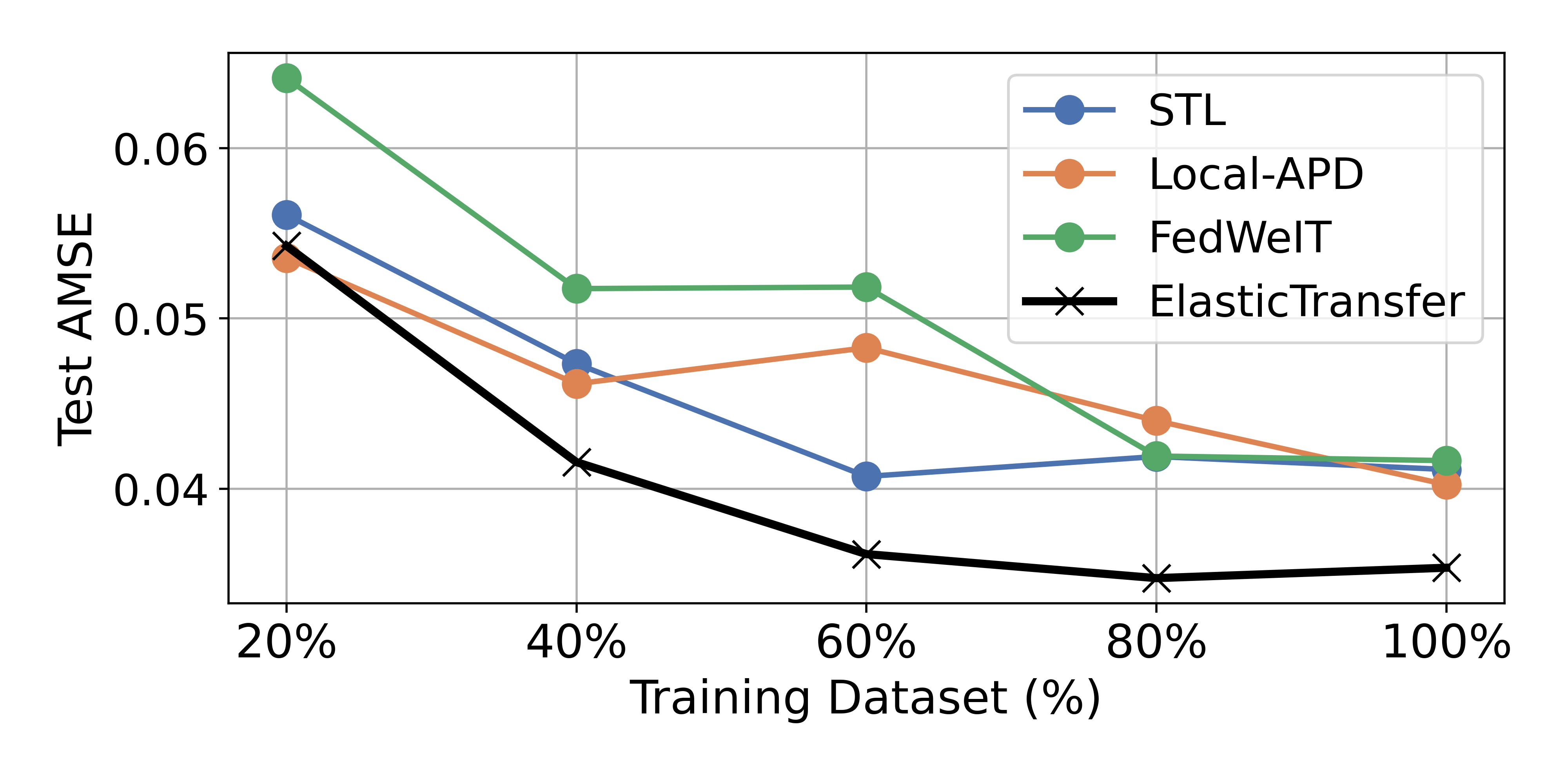}
    \caption[Experiment 2 results.]{AMSE as a function of augmented training dataset size.}
    \label{fig:ablation}
\end{figure}

\subsubsection{Baseline Approaches}

We compare Elastic Transfer with the set of approaches evaluated by \cite{yoonFCL2020} in an image recognition setting. Single Task Learning (STL) methods train a separate model on each local client-task dataset. Local-CL methods (reported in experiments as Local-$\langle$CL method$\rangle$) train a separate model on each client using CL and do not share parameters via FL. FL-CL approaches (reported in experiments as $\langle$FL method$\rangle$-$\langle$CL method$\rangle$) enable clients to communicate using FL as they solve their local CL problem. We consider both architectural and regularization based strategies for FL (FedProx \cite{li2018fedprox} and FedCurv \cite{shoham2019fedcurv}) and CL (APD \cite{yoon2019scalable}, EWC \cite{kirkpatrick2017overcoming}, and L2-transfer), in addition to the FedWeIT algorithm proposed in \cite{yoonFCL2020}. 

\begin{figure}
\centering
\includegraphics[width=\linewidth]{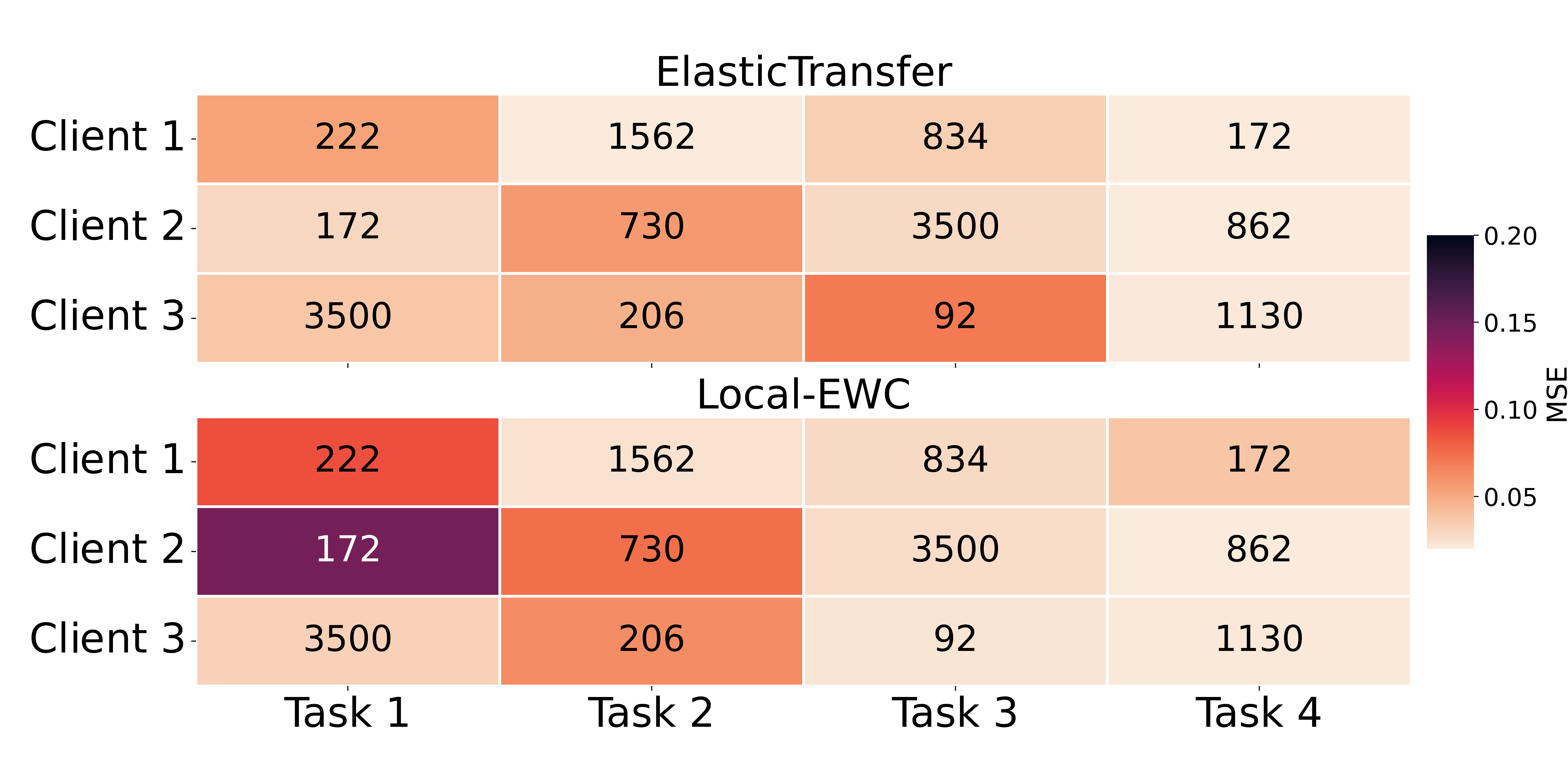}
\caption{Breakdown of client-task partition MSE. Dataset size shown in rectangles.}
\label{fig:client-task}
\end{figure}

\subsection{Experiments and Results}

Our first experiment compares Elastic Transfer with FedWeIT and other baselines evaluated by \cite{yoonFCL2020}. We perform $R=25$ rounds of local training and FL parameter aggregation per task, with $E=5$ local epochs of SGD per FL round. To simulate intermittent robot connectivity, we randomly drop one of the three clients from training each round.

Table \ref{table:exp1} shows results of our first experiment. Elastic Transfer out-performs other methods including FedWeIT, and attains an AMSE that is comparable to our centralized benchmark MSE (0.0363 vs. 0.0296; 22.6 \% increase). We report Elastic Transfer performance over different hyperparamter settings in Table \ref{table:exp3_extended}. Further, other FL-CL approaches out-perform Local-CL approaches that do not perform FL communication. We measure how much task-specific performance deteriorates after task-specific data becomes unavailable during training on later tasks using BWT \cite{lesrotCLRobots}. Elastic Transfer also out-performs baselines in this approach, indicating it suffers from less inter-client interference during training. We also measure adaptation time via FWT. Elastic Transfer demonstrates the best generalization to unseen participants as measured by FWT, followed by other FL-CL approaches. To assess one aspect of the overhead criteria, we \lgdelete{also }report how many static parameters (used for regularization) and trainable parameters (learned via optimization) are required by each method. Compared to FedWeIT, Elastic Transfer requires fewer trainable parameters, but more static parameters. 

Approaches that require learning task specific parameters perform poorly in our evaluation (e.g., FedWeIT, APD). We hypothesize that this is because learning a task-specific parameter decomposition is challenging in low-data settings. Therefore, we vary the size of the training data available in local client-task partitions. If a method demonstrates low AMSE when little data is available, this suggests that the approach \lgedit{provides} improved \textit{adaptation quality} by learning more efficiently. In this experiment, we set $R=60$, $E=1$, with no clients dropped each round. Figure \ref{fig:ablation} shows AMSE as a function of dataset size for several approaches. As training data increases, the relative AMSE gain offered by Elastic Transfer increases. This suggests that Elastic Transfer is leveraging knowledge sharing among client-task datasets effectively as data increases. However, all approaches perform similarly in the lowest data condition (20\%).

FL-CL methods such as Elastic Transfer may perform well because they more effectively share knowledge among client-task datasets. Figure \ref{fig:client-task} demonstrates this in practice by comparing client-task dataset MSE among a FL-CL method (Elastic Transfer) and a Local-CL method (Local-EWC). Local-EWC performs poorly on small client-task datasets, particularly at the beginning of training (e.g., client 2, task 1). This is because Local-CL methods rely on data available to the robot, and do not share knowledge with other robots. Conversely, FedAvg-EWC performs better on these client-task datasets because it leverages knowledge from other clients (e.g., client 3, task 1) to accelerate learning. This shows that FL-CL methods better-facilitate \textit{knowledge sharing} and thus improve \textit{adaptation time} in low-data settings. 

\begin{table}
\begin{tabular}{p{1.8cm}p{.8cm}p{.8cm}p{.8cm}p{.9cm}p{.9cm}} 
\toprule
\textbf{\multirow{2}{*}{\textbf{}}}  & \multicolumn{3}{c}{Evaluation Metrics~} & \multicolumn{2}{c}{Parameters~}  \\ 
     &   AMSE   &  BWT  &  FWT  & Static & Trainable \\
\midrule
\textbf{Local~} &          &          &          &           &          \\
\midrule
STL             &  0.0735 &  0.0332 &  0.5388 &       0 &  30,724 \\
\midrule
\textbf{Local-CL~} &          &          &          &           &   \\
\midrule
Local-EWC       &  0.0482 &  0.0272 &  0.0641 &   61,448 &   7,681 \\
Local-APD       &  0.0717 &  0.1041 &  0.1766 &       0 &  39,176 \\
\midrule
\textbf{FL-CL~} &          &          &          &           &          \\
\midrule
FedAvg-SGD      &  0.0391 &  0.0088 &  0.0369 &       0 &   7,681 \\
FedProx-SGD     &  0.0408 &  0.0093 &  0.0371 &    7,681 &   7,681 \\
FedAvg-EWC  & 0.0386 & 0.0128 & 0.0373 & 61,448 & 7,681 \\
FedProx-EWC     &  0.0422 &  0.0124 &  0.0470 &   69,129 &   7,681 \\
FedAvg-APD      &  0.0927 &  0.2378 &  0.6182 &       0 &  39,176 \\
FedProx-APD     &  0.0665 &  0.1393 &  0.3187 &    7,424 &  39,176 \\
FedWeIT         &  0.0886 &   0.1597 &  0.5603 &   89,100 &  39,272 \\
\midrule
Elastic Transfer &  \textbf{0.0363} &  \textbf{0.0083} &  \textbf{0.0352} &  115,215 &   7,681 \\

\bottomrule
\end{tabular}

\caption{Results comparing Elastic Transfer with previously-proposed approaches in our FCRL benchmark. All results computed on held-out test data. Table shows Average MSE (AMSE), Backwards Transfer (BWT), Forwards Transfer (FWT).}\label{table:exp1}

\end{table}

\begin{table}[th!]
\centering
\resizebox{\columnwidth}{!}{%
\label{exp3_results_extended}
\begin{tabular}{lll|ccc|cc} 
\toprule
\multicolumn{3}{c|}{\multirow{2}{*}{\textbf{Algorithm}}}   & \multicolumn{3}{c}{\textbf{Evaluation Metrics~}} & \multicolumn{2}{c}{\textbf{Parameters~}} \\ 
    &  &     &   \textbf{AMSE}   &  \textbf{BWT}    &  \textbf{FWT}  & \textbf{S} & \textbf{T} \\
    $\lambda_1$  & $\lambda_2$  & $\lambda_3$ &                 &                 &     &       &    \\ 
    \hline
\multirow{9}{*}{$5e^{-1}$}                   &                                            & $5e^{-1}$       & 0.0408                     & 0.0120                     & 0.0380             &  \multirow{27}{*}{115,215}   & \multirow{27}{*}{7,681}    \\
                                             & $5e^{-1}$                                  & $5e^{-1}$       & 0.0389                     & 0.0108                     & 0.0407             &      &     \\
                                             &                                            & $5e^{-1}$       & 0.0383                     & 0.0107                     & 0.0383             &      &     \\
\cline{2-6}
                                             &                                            & $5e^{-2}$       & 0.0392                     & 0.0114                     & 0.0391             &      &     \\
                                             & $5e^{-2}$                                  & $5e^{-2}$       & 0.0385                     & 0.0115                     & 0.0390             &      &     \\
                                             &                                            & $5e^{-2}$       & 0.0369                     & 0.0124                     & 0.0355             &      &     \\
\cline{2-6}
                                             &                                            & $0$             & 0.0389                     & 0.0103                     & 0.0395             &      &     \\ 
                                             & $0$                                        & $0$             & 0.0397                     & 0.0113                     & 0.0450             &      &     \\ 
                                             &                                            & $0$             & 0.0391                     & 0.0085                     & 0.0362             &      &     \\ 
\cline{1-6}

\multirow{9}{*}{\textbf{\textbf{$5e^{-2}$}}} &                                            & $5e^{-1}$       & 0.0396                     & \textbf{0.0062}            & 0.0400             &      &     \\
                                             & $5e^{-1}$                                  & $5e^{-2}$       & 0.0402                     & 0.0115                     & 0.0381             &      &     \\
                                             &                                            & $0$             & 0.0376                     & 0.0107                     & 0.0375             &      &     \\
\cline{2-6}
                                             &                                            & $5e^{-1}$       & 0.0400                     & 0.0111                     & 0.0395             &      &     \\
                                             & $5e^{-2}$                                  & $5e^{-2}$       & 0.0390                     & 0.0080                     & 0.0392             &      &     \\
                                             &                                            & $0$             & 0.0390                     & 0.0091                     & 0.0391             &      &     \\
\cline{2-6}
                                             &                                            & $5e^{-1}$       & 0.0384                     & 0.0115                     & 0.0390             &      &     \\ 
                                             & $0$                                        & $5e^{-2}$       & 0.0391                     & 0.0086                     & 0.0382             &      &     \\ 
                                             &                                            & $0$             & 0.0386                     & 0.0128                     & 0.0373             &      &     \\ 
\cline{1-6}

\multirow{9}{*}{0}                           &                                            & $5e^{-1}$       & 0.0403                     & 0.0112                     & 0.0412             &      &     \\
                                             & $5e^{-1}$                                  & $5e^{-2}$       & 0.0388                     & 0.0122                     & 0.0396             &      &     \\
                                             &                                            & $0$             & \textbf{0.0363}            & 0.0083                     & \textbf{0.0352}    &      &     \\
\cline{2-6}
                                             &                                            & $5e^{-1}$       & 0.0392                     & 0.0118                     & 0.0380             &      &     \\
                                             & $5e^{-2}$                                  & $5e^{-2}$       & 0.0374                     & 0.0108                     & 0.0384             &      &     \\
                                             &                                            & $0$             & 0.0374                     & 0.0101                     & 0.0358             &      &     \\
\cline{2-6}
                                             &                                            & $5e^{-1}$       & 0.0378                     & 0.0092                     & 0.0382             &      &     \\ 
                                             & $0$                                        & $5e^{-2}$       & 0.0404                     & 0.0137                     & 0.0411             &      &     \\ 
                                             &                                            & $0$             & 0.0388                     & 0.0122                     & 0.0355             &      &     \\
\bottomrule
\end{tabular}
}

\caption[Experiment 1 full results.]{Results of full set of Elastic Transfer hyperparameter combinations.}\label{table:exp3_extended}
\end{table}

\section{Conclusion and Future Work}

In this paper, we presented a new decentralized learning framework for socially aware robotics (FCL4SR), by capitalising on two machine learning approaches, Federated Learning and Continual Learning, to capture interaction dynamics distributed physically across robots and temporally across repeated robot encounters. We outlined four criteria (adaptation quality, adaptation time, knowledge sharing, and model overhead) that should be balanced within the FCL4SR framework. We also developed a new regularization-based FCL method -- Elastic Transfer, and evaluated it on a proof-of-concept socially aware navigation dataset, demonstrating that Elastic Transfer improved adaptation quality, adaptation time, and knowledge sharing. 

\lgedit{Our FCL4SR framework (i.e., Elastic Transfer) only requires temporary local storage of end-user data during model training, but requires no permanent storage of end-user data on a server. Therefore, our approach provides an important first step towards end-user privacy protection compared to current centralized learning practices involving permanent data storage. However, our approach does \textit{not} provide a formal privacy guarantee \cite{mcmahan2017fedavg}. Future work could investigate how our framework might be combined with other privacy-protecting methods, such as differential privacy.} \lgdelete{There are several opportunities to expand this work. FCL is an extensible decentralized learning framework that can be applied in other socially aware robotics and human robot interaction settings. Based on the promising results provided by Elastic Transfer, future work can focus on extending this approach to new machine learning domains, and evaluate it thoroughly on a series of broader benchmarks.} We also see opportunities \lgedit{for future work} to further reduce communication overhead by leveraging similar reformulations introduced by FedCurv and Online-EWC. We hope that this work will serve as a stepping stone for improving the personalization capabilities of socially aware robots while also paving the way towards privacy preservation in human robot interactions. 
\lgedit{
\section*{ACKNOWLEDGMENTS}
\footnotesize
\textbf{Funding:} This work is partially supported by the EPSRC/UKRI under grant ref. EP/R030782/1 (ARoEQ). \textbf{Open Access:} For open access purposes, the authors have applied a Creative Commons Attribution (CC BY) licence to any Author Accepted Manuscript version arising.}
\bibliographystyle{IEEEtran}
\bibliography{refs}  

\end{document}